\documentclass[letterpaper]{article} 
\usepackage{aaai24}  
\usepackage{times}  
\usepackage{helvet}  
\usepackage{courier}  
\usepackage[hyphens]{url}  
\usepackage{graphicx} 

\usepackage{multirow}
\usepackage{array}
\usepackage{caption}
\usepackage{booktabs}

\usepackage{natbib}  
\usepackage{caption} 
\usepackage{algorithm}
\usepackage{algorithmic}
\usepackage{amsfonts}

\usepackage{newfloat}
\usepackage{listings}
\DeclareCaptionStyle{ruled}{labelfont=normalfont,labelsep=colon,strut=off} 
\floatstyle{ruled}
\newfloat{listing}{tb}{lst}{}
\floatname{listing}{Listing}
\title{ExpCLIP: Bridging Text and Facial Expressions via Semantic Alignment}
\author{
    Yicheng Zhong\equalcontrib,
    Huawei Wei\equalcontrib,
    Peiji Yang\equalcontrib,
    Zhisheng Wang
}
\affiliations{
    \textsuperscript{\rm 1}  Tencent Technology (Shenzhen) Co.Ltd

    \{ajaxzhong, huaweiwei, peijiyang, plorywang\}@tencent.com
}

\usepackage{bibentry}

\begin{document}

\twocolumn[{%
\renewcommand\twocolumn[1][]{#1}%
\maketitle
\begin{center}
    \centering
    \includegraphics[width=0.8\textwidth]{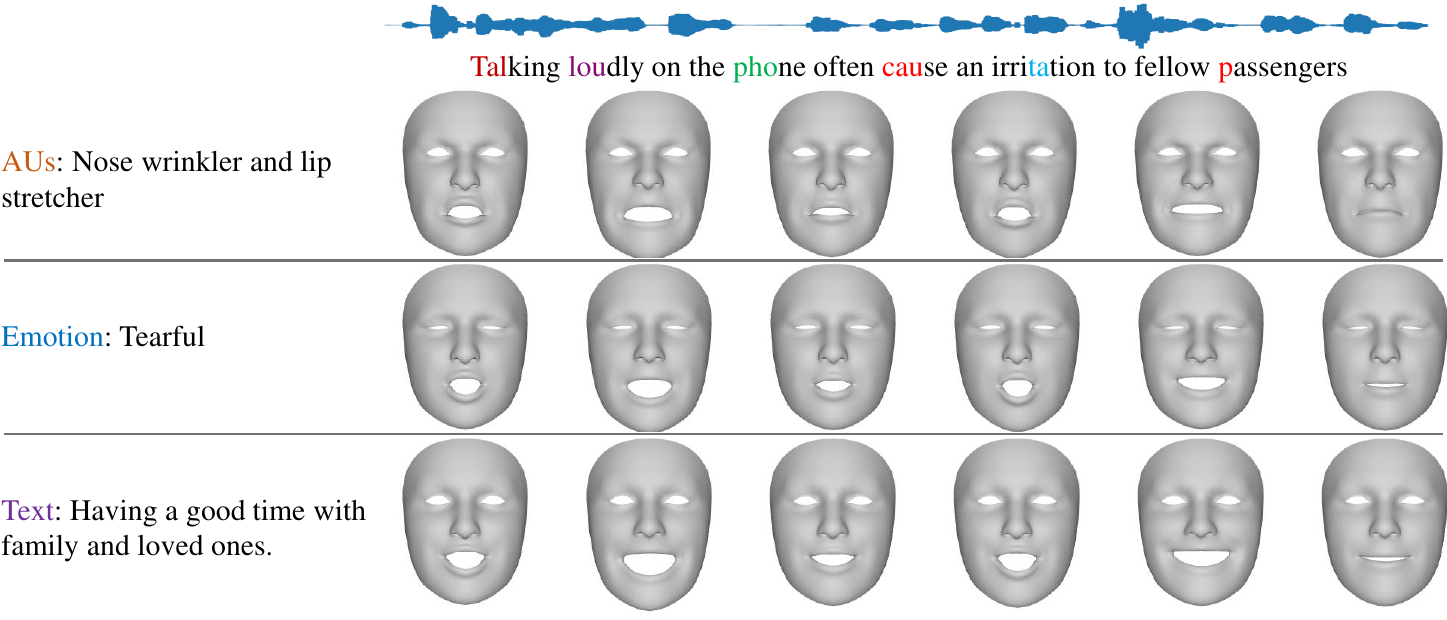}
    \captionof{figure}{Illustration of our text-guided emotional speech-driven facial animation generation, our approach accommodates diverse textual inputs for style control, encompassing AUs, emotion tags, and other forms of natural language.}
    \label{fig:first result}
    \end{center}%
}]

\begin{abstract}


The objective of stylized speech-driven facial animation is to create animations that encapsulate specific emotional expressions. Existing methods often depend on pre-established emotional labels or facial expression templates, which may limit the necessary flexibility for accurately conveying user intent.
In this research, we introduce a technique that enables the control of arbitrary styles by leveraging natural language as emotion prompts. This technique presents benefits in terms of both flexibility and user-friendliness.
To realize this objective, we initially construct a Text-Expression Alignment Dataset (TEAD), wherein each facial expression is paired with several prompt-like descriptions. We propose an innovative automatic annotation method, supported by Large Language Models (LLMs), to expedite the dataset construction, thereby eliminating the substantial expense of manual annotation.
Following this, we utilize TEAD to train a CLIP-based model, termed ExpCLIP, which encodes text and facial expressions into semantically aligned style embeddings. The embeddings are subsequently integrated into the facial animation generator to yield expressive and controllable facial animations. Given the limited diversity of facial emotions in existing speech-driven facial animation training data, we further introduce an effective Expression Prompt Augmentation (EPA) mechanism to enable the animation generator to support unprecedented richness in style control.
Comprehensive experiments illustrate that our method accomplishes expressive facial animation generation and offers enhanced flexibility in effectively conveying the desired style.
\end{abstract}

\section{Introduction}

In recent years, speech-driven facial animation has gained importance due to its widespread applications in diverse fields such as gaming, virtual reality, and film production \cite{zhen2023human}.
Currently, most research focuses on improving the synchronization between lip movements and speech \cite{Cudeiro2019CaptureLA,Fan2021FaceFormerS3,chen2022transformer,xing2023codetalker}. This emphasis only allows for conveying speech content, not style, resulting in a lack of emotional expressions in generated facial animations.
A few works \cite{karras2017audio, danvevcek2023emotional} have attempted to integrate emotions into facial animation by providing the model with specific emotional labels or use reference facial expressions as style guidance. However, they either have limited flexibility in expressing diverse emotions or necessitate searching for a reference image or video, which may be impractical for users.

In this study, we propose to adopt natural language as the style prompt for emotional facial animation generation, which offers both flexibility and user-friendliness. 
A straightforward approach is to collect animation data with paired text prompts and train a text-guided animation generator. However, the scarcity of such data and the high cost of annotation make this approach infeasible.
To address this challenge, we propose a novel CLIP-based model called ExpCLIP in this paper. ExpCLIP is designed to learn an embedding space where the representations of text and facial expressions are semantically aligned. By leveraging the capabilities of ExpCLIP, we can train the animation generator by specifying representative facial expressions as prompts, which can be easily extracted from animations, and utilize text prompts for inference purposes.

To train ExpCLIP, a large-scale text-expression dataset is required. However, currently available datasets \cite{wang2020mead,kollias2022abaw} only have limited tag-level emotion labels. To address this issue, we propose a novel automated annotation method to construct a Text-Expression Aligned Dataset (TEAD). Specifically, we leverage the visual understanding  capability of large language models (LLMs) \cite{radford2019language,brown2020language,ouyang2022training} to accomplish the annotation task. We find that LLMs are capable of describing the facial expressions corresponding to an emotional text.
By harnessing the power of LLMs, we collect a rich emotional corpus and use meticulously engineered prompts to ask the LLMs to output the corresponding description of facial expressions. Here, we use activated facial Action Units (AUs) to describe facial expressions.

Building upon ExpCLIP, which is trained on TEAD, we propose an emotion-controllable facial animation generator.
During the training phase of the generator, we employ a self-attention module to extract the expression prompt from an animation clip. 
Subsequently, the expression prompt is fed into ExpCLIP to obtain the emotion embedding, which is then fused with the input speech to generate the target facial animation. During the inference stage, we can use text prompts to achieve the desired style control, which is attributed to the alignment of facial expressions and text embeddings achieved by ExpCLIP. We show several examples in Figure \ref{fig:first result}.
Notably, our framework can be readily extended to support facial images as input prompts. This can be achieved by  augmenting  ExpCLIP with an image encoder and training it on paired facial images and facial expressions.  The paired data can be easily obtained with monocular 3D face reconstruction methods \cite{guo2022perspective,lei2023hierarchical,chai2023hiface}.


Moreover, we propose an effective Expression Prompt Augmentation(EPA) mechanism to enable the animation generator to handle unseen emotions. This mechanism involves incorporating random perturbations to the expression prompt and devising a lip motion constraint to ensure the generated lip motions remain consistent with the original motions. The underlying assumption is that when individuals articulate the same sentence with different emotions, their lip motions tend to be consistent.

In summary, the main contributions of our research are:
\begin{itemize}
    \item We leverage the visual understanding  capability of LLMs to propose an automatic annotation method for inferring facial expressions from emotional text. This enables us to construct a large-scale text-expression aligned dataset. 
    \item We propose ExpCLIP, which is capable of aligning the semantic representations of text and facial expressions, empowering the inference of emotional styles from natural language descriptions.
    \item We present the first attempt to use natural language text as prompts to achieve flexible and controllable emotional speech-driven facial animation generation. 
    
\end{itemize}

\section{Related Work}
\subsection{Speech-driven face animation}
This field has witnessed a surge of substantial efforts in recent years. VOCA \cite{Cudeiro2019CaptureLA} interprets the mapping from speech to animation as a regression problem. FaceFormer \cite{Fan2021FaceFormerS3} leverages transformers to capture the long-term dependencies inherent in speech. Meshtalk \cite{Richard2021MeshTalk3F} prioritizes addressing the model's scalability and realism by segregating audio-correlated and audio-uncorrelated information. However, all these methods primarily focus on enhancing lip synchronization, falling short in conveying emotional expressions.

\subsection{Emotion guided generation}
The incorporation of emotional expressions has been considered in recent works. \cite{karras2017audio} uses a trainable variable to represent the emotion state and \cite{danvevcek2023emotional} uses predefined emotion labels to guide the animation generation. Similar approaches have also been observed in several works\cite{sadoughi2019speech,ji2021audio,wu2021imitating,liang2022expressive,sinha2022emotion} related to talking face generation. \cite{ji2022eamm} and \cite{ma2023styletalk} propose to extract emotional information from reference videos. \cite{wang2023progressive} utilize a static facial image as an emotional condition for styled talking face generation. These methods either have limited flexibility in expressing diverse emotions, or necessitate searching for a reference image or video, which may be impractical for users.

\subsection{CLIP-based content synthesis}
CLIP \cite{radford2021learning} has demonstrated its efficacy text guided image editing \cite{rombach2022high,ramesh2023hierarchical, schaldenbrand2021styleclipdraw}. It can also be extended to the integration of text with other modalities, such as employing CLIP for 3D motion generation \cite{tevet2022motionclip}. However, to the best of our knowledge, no prior work has utilized CLIP for 3D facial animation generation. In this paper, we propose, for the first time, the use of CLIP for text-guided speech-driven facial animation generation.

\begin{figure*}[!htbp]
    \centering
    \includegraphics[width=0.9\linewidth]{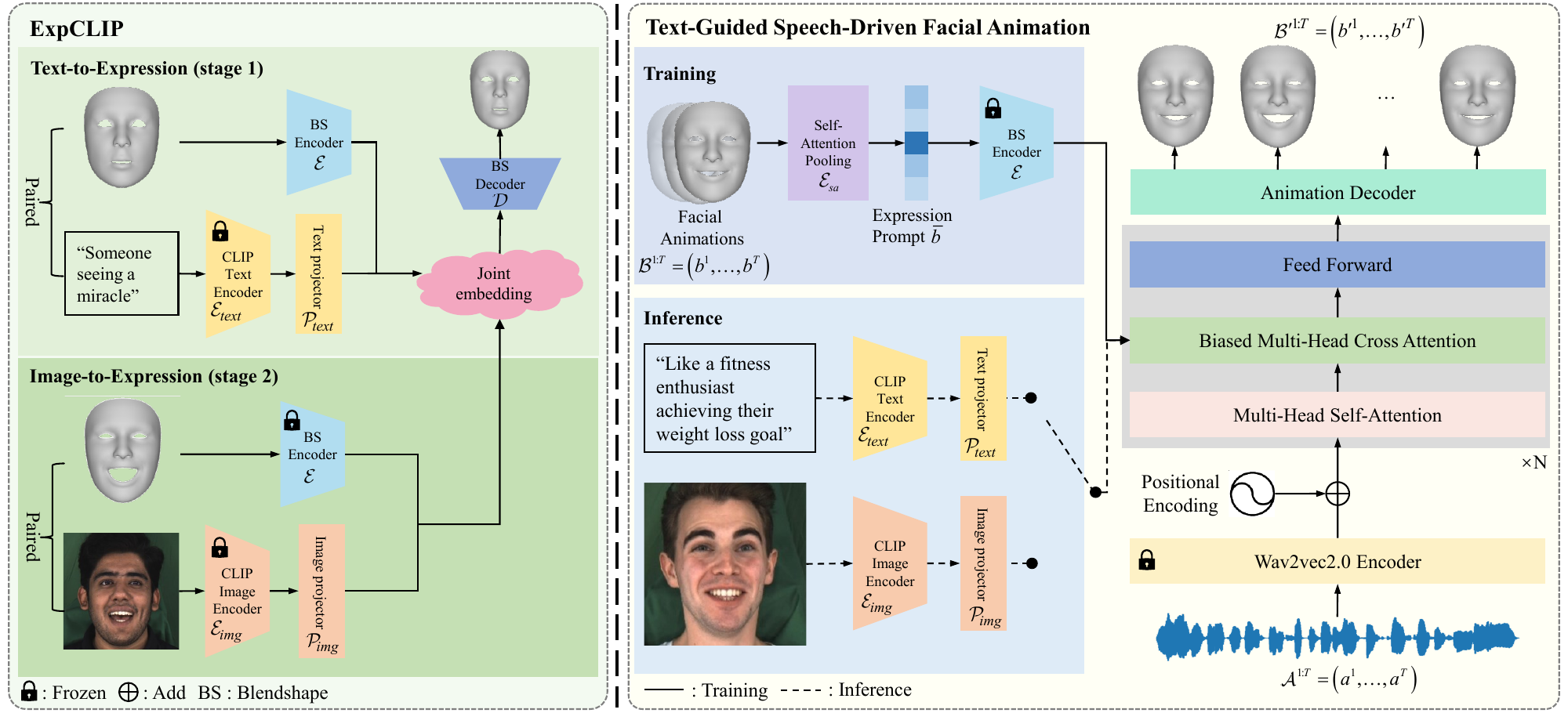}
    \caption{Overview of our framework. We first train ExpCLIP to establish semantic alignment between text ,facial expressions and facial images. Then, we employ it for the generation of emotional speech-driven facial animation. The animation generator supports both images and natural language prompts for emotion control. }
    \vspace{-0.2cm}
    \label{fig:overview}
\end{figure*}

\section{Method}
In this section, we first present the construction of TEAD, followed by an overview of the training process for ExpCLIP. Finally, we introduce the proposed text-guided speech-driven facial animation method.

\subsection{Text-expression aligned dataset} 

In order to train a model capable of aligning the semantic representations of natural language and facial expressions, a text-expression dataset is
required. However, the currently available datasets only provide limited  emotion labels \cite{cao2014cremad,wang2020mead}, which are insufficient for achieving the desired alignment at a fine-grained level. To address this limitation, we propose a Text-Expression Aligned Dataset (TEAD), which is automatically constructed with the assistance of LLMs.

\subsubsection{Motivation} 

To generate paired emotional text and corresponding facial expression data using LLMs, it is imperative to represent facial expressions as textual descriptions. Fortunately, the Facial Action Coding System (FACS) \cite{ekman1978facial} offers a systematic approach to describe human facial movements by decomposing facial expressions into independent AUs, with each AU possessing an exhaustive textual description.  
Therefore, our objective is to let LLMs generate corresponding activated AUs from the emotional text.
This necessitates that LLMs demonstrate cross-modal understanding capability, that is, the ability to “\textbf{imagine}” corresponding facial expressions that depict the emotion in the text, and subsequently translate the expressions into activated AUs. Through our testing, we find that LLMs can successfully accomplish this task with a carefully designed prompt through in-context learning. 

\subsubsection{Automatic data generation} 
\begin{figure}
    \centering
\includegraphics[width=0.9\linewidth]{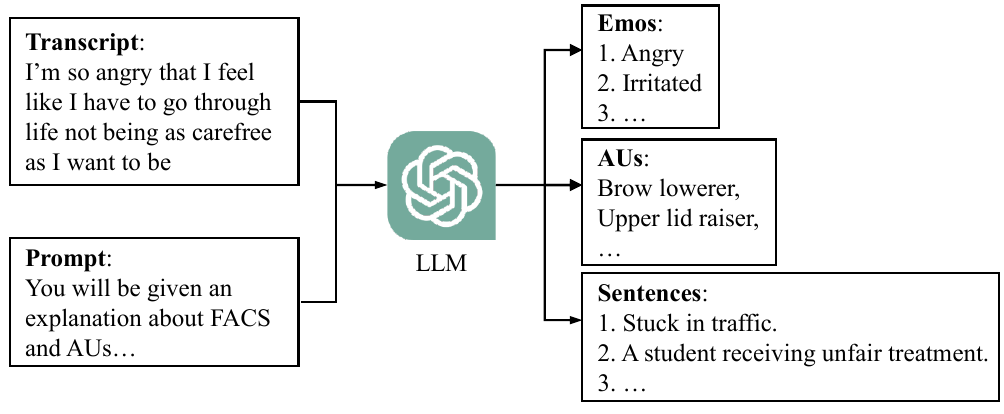}
    \caption{An example of data generation of TEAD.}
    \label{fig:tead}
    \vspace{-0.2cm}
\end{figure}

We utilize the abundant corpus from text emotion classification tasks \cite{MohammadB17starsem}, which encompass rich real-world human emotions. The emotional transcripts are fed into the LLM for text emotion classification and detection of activated AUs.  Specifically, given an emotional transcript $t$, the LLM predicts its emotional tags $e$, where $e$ contains several emotional labels, usually 3 to 5, and the activated AUs with a one-hot vector $u=\{u_i, i=1,...,N_u\}$. $N_u$ is 36 in our work. Note the combination of these AUs enables a wider range of emotions compared to existing datasets that have only a few emotion labels. We further prompt the LLM to describe situations that may evoke inferred emotions, generating sentence-level labels $s$. An instance is shown in Figure \ref{fig:tead}. Detailed information of our engineered prompts can be found in the Supplementary Materials.

In addition, we enlist a professional facial animator to devise a mapping rule to convert AUs into blendshape weights. Then, we transform the AUs vector $u$ into a set of blendshape weights $b \in \mathbb{R}^{52}$, enabling us to leverage publicly available datasets that represent facial expressions as blendshape weights. Consequently, TEAD can be represented as a set of quadruples:$\mathcal{T} = \{(t, e, b, s)_i| i=1,...,N_T\}$.  About 50,000 quadruples are included.

\subsection{ExpCLIP} 

We utilize TEAD to train a CLIP-based model, named ExpCLIP.  As illustrated in Figure \ref{fig:overview}, ExpCLIP is an autoencoder framework designed to align multimodal signals, encompassing diverse human facial expressions, textual descriptions of emotions, and realistic facial images.

\subsubsection{Text-to-expression} 


We propose a blendshape encoder $\mathcal{E}$ that maps a given set of blendshape weights $b$ into an embedding. Subsequently, a decoder $\mathcal{D}$ reconstructs the blendshape weights  from the embedding.  In parallel, we utilize a CLIP text encoder $\mathcal{E}_{text}$  based on the work of \cite{radford2021learning},  along with a text projector $\mathcal{P}_{text}$ to map the  emotion text, sampled from \{$t$, $e$, $s$\}, into the joint embedding space. Due to the strong generalization capability of pre-trained CLIP, we keep $\mathcal{E}_{text}$ frozen during training to leverage its acquired extensive knowledge.

\subsubsection{Image-to-expression} 
We propose an extension to incorporate facial images as emotion prompts by integrating an image encoder into ExpCLIP and aligning the embeddings of images and expressions. A SOTA 3D face reconstruction method \cite{guo2022perspective} is used to create paired blendshape weights for facial images. We leverage the pre-trained image encoder $\mathcal{E}_{img}$  from CLIP and employ an image projector $\mathcal{P}_{img}$ to map facial images into the joint embedding space. During this step, we only finetune $\mathcal{P}_{img}$, while keeping the the weights of $\mathcal{E}_{img}$, $\mathcal{E}$, and $\mathcal{D}$  fixed.

\subsubsection{Objective function} 
ExpCLIP is trained via three types of losses: auto-encoder reconstruction loss, embedding alignment loss, and cross-modal reconstruction loss:
\begin{equation}
\label{eq:loss_t_all}
    \mathcal{L} = \lambda_{1} \mathcal{L}_{ae} + \lambda_{2} \mathcal{L}_{emb}^{d} + \lambda_{3} \mathcal{L}_{cross}^{d},
\end{equation}
where $d \in \{text, image\}$. Note that $d=text$ indicates the text-to-expression training, and $d=image$ refers to the image-to-expression finetuning. The auto-encoder reconstruction loss $\mathcal{L}_{ae}$ measures the L2 distance between the input blendshape weights $b$ and the predicted ones:
\begin{equation}
    \mathcal{L}_{ae} = \mathbb{E}_{b\sim \mathcal{T}} \Vert \mathcal{D}(\mathcal{E}(b)) - b \Vert.
\end{equation}

We align the embeddings of text (or images) and expressions using cosine embedding loss:
\begin{equation}
\label{eq:loss_t_la}
    \mathcal{L}_{emb}^{d} = 1 - {\rm cos} (\mathcal{P}_{d}(\mathcal{E}_{d} (S)) - \mathcal{E}(b)),
\end{equation}
where $S$ represents text in $(t, e, s)$ when $d=text$, and facial images when $d=image$.

To better improve the multi-modal alignment, we propose a cross-modal reconstruction loss, which enforces the blendshape weights reconstructed using text (or image) embeddings close to its paired ones, namely:
\begin{equation}
\label{eq:loss_t_cr}
    \mathcal{L}_{cross}^{d} = \mathbb{E}_{b\sim \mathcal{T}} \Vert \mathcal{D}(\mathcal{P}_{d}(\mathcal{E}_{d}(S))) - b \Vert.
\end{equation}

We propose several strategies to enhance the model's generality. They include fully exploiting the TEAD dataset by randomly extracting samples from $t$, $e$, and $s$ as text inputs for training. Additionally, we  borrow text augmentation techniques from the natural language processing literature \cite{wei2019eda}, such as stop-word removal, synonym replacement, and sentence shuffling. Furthermore, we augment the blendshape weights by applying minor random perturbations, which do not alter the emotions expressed by the weights.  Experiments show all these strategies improve the model's robustness.


\subsection{Text-guided speech-driven facial animation}
Our aim is to automatically generate emotionally expressive facial animation, where the content is determined by speech and the emotional style is controlled by a text prompt. The overview of our method is shown in Figure \ref{fig:overview}.
\subsubsection{Training Workflow}

Leveraging the semantic alignment between text and facial expressions achieved by ExpCLIP, we can employ expression prompts for training and text prompts for inference. Yet, annotating expression prompts for each animation clip is costly. To address this, we propose a self-attention mechanism to automatically extract representative expressions from the animation clips. Consequently, the training workflow of our model is as follows:

Let  $\mathcal{A}^{1: T}=\left({a}^{1}, \ldots, {a}^{T}\right)$ denotes a sequence of speech snippets. $\mathcal{B}^{1: T}=\left({b}^{1}, \ldots, {b}^{T}\right)$ is the synchronized facial animation, each frame is represented by a set of blendshape weights ${b}^{t}$. 
We propose a transformer-based self-attention pooling module, denoted as $\mathcal{E}_{sa}$, to derive the  attention weights for individual frames within the animation clip. Subsequently, we aggregate all frames of the clip according to the attention weights, resulting in the generation of the expression prompt $\overline{b}$. 
Next, we feed the expression prompt  $\overline{b}$ into the expression encoder $\mathcal{E}$ of ExpCLIP to obtain the style embedding. During the training process, the parameters of $\mathcal{E}$ are kept frozen. Simultaneously, we utilize a pre-trained wav2vec2.0 model \cite{baevski2020wav2vec} to convert raw waveform input into contextualized speech features. We employ a transformer decoder to predict $\mathcal{B'}^{1: T}=\left({b'}^{1}, \ldots, {b'}^{T}\right)$ from speech features, with style embedding incorporated into the decoding process via cross-attention. 

A simple L1 loss is utilized for reconstruction:
\begin{equation}
    \mathcal{L}_{rec} = \frac{1}{T}\sum ^{T}_{t=0} | {b'}^{t} - {b}^{t} |
\end{equation}

\subsubsection{Expression Prompt Augmentation}
The currently publicly accessible  speech-driven facial animation datasets exhibit insufficiency in terms of emotional richness. They are typically constrained to a limited number of coarse-grained emotion labels, thereby impeding the model's ability to handle unseen fine-grained emotions. To address this issue, we propose an Expression Prompt Augmentation (EPA) mechanism. Specifically, we add perturbations to the expression prompts. In order to make the perturbed prompts show certain emotions instead of random weird expressions, the perturbations are obtained by randomly sampling a facial expression $b_{aug}$ from TEAD. Subsequently, we blend $b_{aug}$ into the original expression prompt using a random weight $\lambda$. 
\begin{equation}
    \overline{b}_{aug} = (1-\lambda) \overline{b} + \lambda b_{aug}, \lambda \in [0, 1]
\end{equation}

Due to the lack of corresponding animations for the perturbed expression prompts, we devise two loss functions to ensure that the generated animations  exhibit accurate lip movements and emotional styles. We posit that when a person utters the same sentence with different emotions, their lip movements remain fundamentally consistent, meaning that the displacement between adjacent frames remains consistent. Leveraging this assumption, we  formulate a lip motion loss to capture and enforce this consistency in the generated animations. We represent the animation for the perturbed prompt as $\mathcal{B'}^{1: T}_{aug}=\left({b'}^{1}_{aug}, \ldots, {b'}^{T}_{aug}\right)$, and the lip motion loss is defined as:
\begin{equation}
    \mathcal{L}_{lm} = \frac{1}{T-1}\sum ^{T-1}_{t=0} | ({b'}^{t+1}_{aug} - {b'}^{t}_{aug}) - ({b}^{t+1} - {b}^{t}) |
\end{equation}

$\mathcal{L}_{lm}$ only ensures the correctness of lip motion in animations generated based on perturbed prompts, but it does not guarantee the desired style. 
To address this limitation, we propose a style loss:
\begin{equation}
    \mathcal{L}_{style} = \Vert \mathcal{E}(\mathcal{E}_{sa}(\mathcal{B'}_{aug}))  - \mathcal{E}(\overline{b}_{aug}) \Vert
\end{equation}
$\mathcal{L}_{style} $ ensures the consistency between the emotion of the generated animation $\mathcal{B'}_{aug}$ and the augmented expression prompt $\overline{b}_{aug}$.


\subsubsection{Inference Phase}
Since the embeddings of text and facial expressions are semantically aligned by ExpCLIP, we can employ text descriptions as prompts to control the emotions of the generated animations. 
There is no strict requirement for the text to adhere to a specific format. You have the freedom to express your text prompts in any manner you prefer. For instance, you may provide a precise description of the desired facial expression or simply describe the mood you want to express. Furthermore, in cases where verbal text may not effectively convey the intended emotions, the option of employing a reference facial image as a prompt is also available. This capability  is facilitated by ExpCLIP's alignment of facial images and facial expressions.

\begin{figure}[t]
    \centering
    \includegraphics[width=0.9\linewidth]{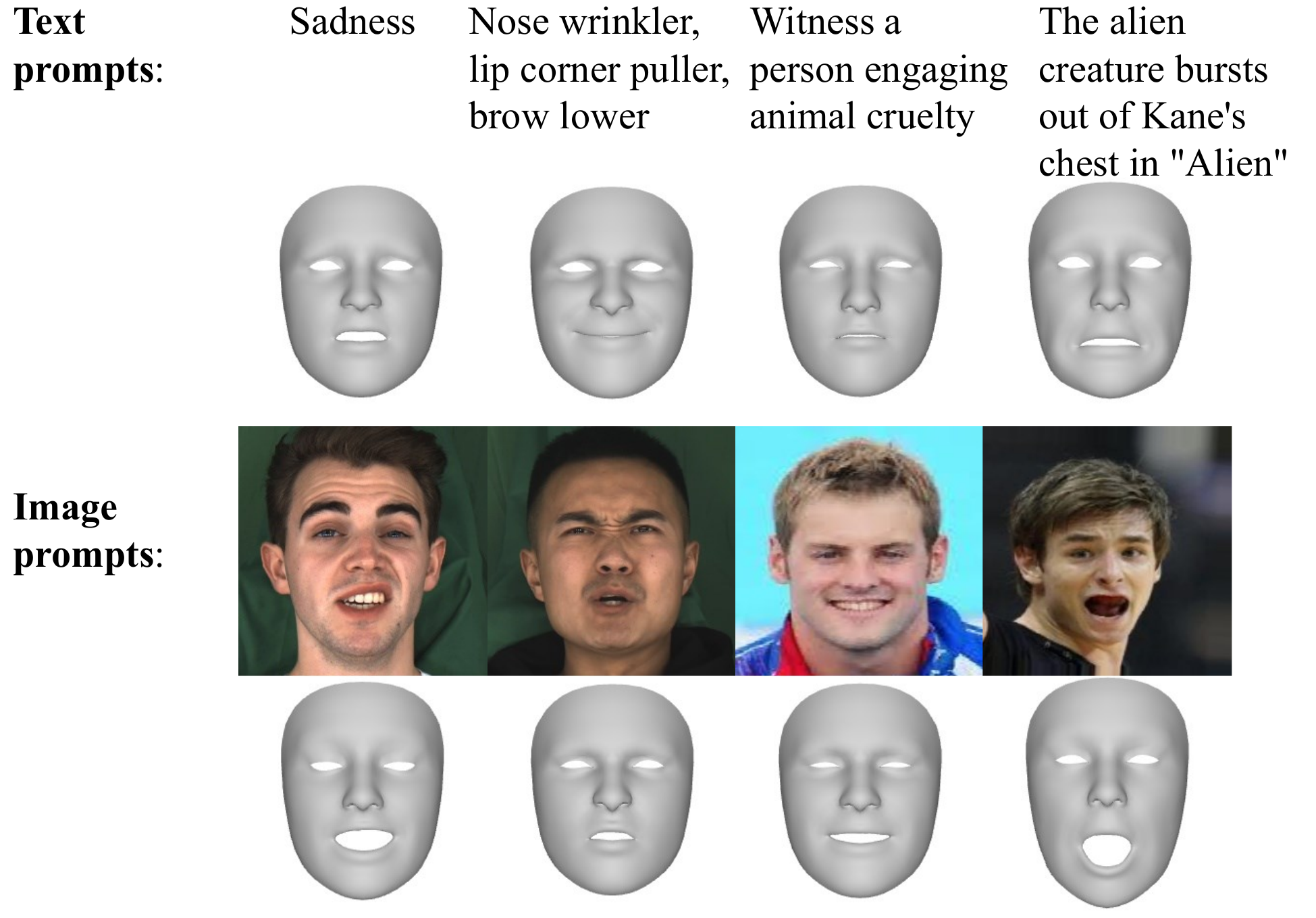}
    \caption{Multi-modal alignment. Row 1\&2: text-to-expression. Row 3\&4: image-to-expression. Note the last two columns are \emph{out-of-domain} samples.}
    \label{fig:text bs}
    \vspace{-0.3cm}
\end{figure}

\begin{figure}[t]
    \centering
    \includegraphics[width=0.9\linewidth]{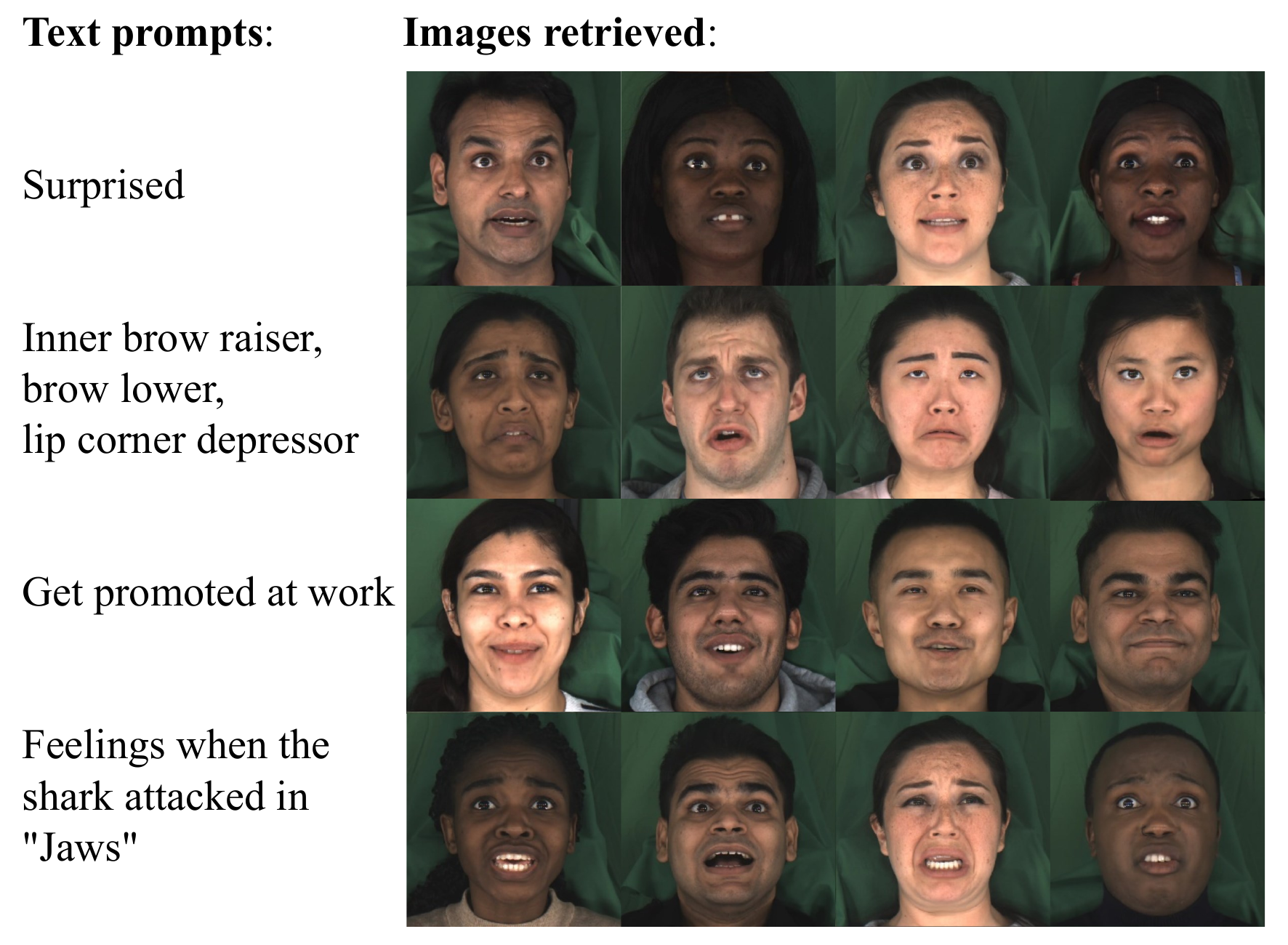}
    \caption{Text-to-image retrieval on MEAD-3D.}
    \label{fig:text img}
    \vspace{-0.3cm}
\end{figure}

\begin{figure}[t]
    \centering
    \includegraphics[width=0.9\linewidth]{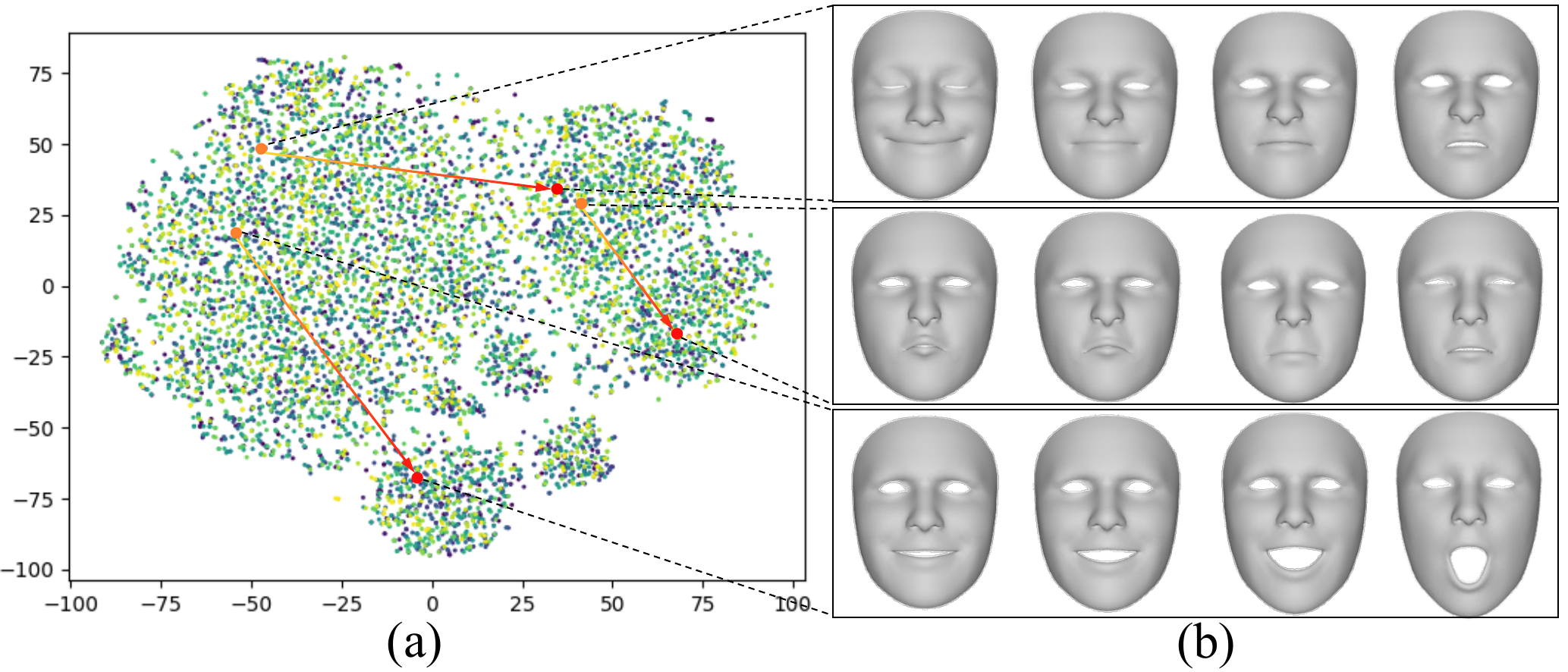}
    \caption{(a) t-SNE of expression embeddings in TEAD. (b) smooth interpolations between two distant expressions. } 
    \vspace{-0.3cm}
    \label{fig:tsne}
\end{figure}

\section{Experiments}
In this section, we first introduce the used datasets and implementation details. Subsequently, we present the capabilities of ExpCLIP in multimodal alignment and discuss several key factors for training ExpCLIP. Finally, we show the promising results of style-controllable speech-driven facial animation, accompanied by detailed comparative experiments and ablation studies.

\subsection{Datasets}
\subsubsection{TEAD}
We train ExpCLIP using the proposed TEAD, which consists of 50,000 quadruples. Each quadruple includes text, a set of emotion tags, AUs, blendshape weights, and situation sentences.
We use 90\% of the data for training and the remaining 10\% for testing the text-expression alignment of ExpCLIP.

\subsubsection{MEAD-3D}
To support the image-expression alignment of ExpCLIP, we generate image-expression paired data based on MEAD . 
MEAD is a talking-face video corpus featuring 60 actors talking with 8 different emotions at 3 different intensity levels. We sample 150,000 images from MEAD and use a SOTA 3D monocular face reconstruction method\cite{guo2022perspective} to obtain the 3D facial meshes. Then we compute the blendshape weights corresponding to each mesh using the SLSQP solver from SciPy\cite{virtanen2020scipy}. We use 90\% of the data for training and the remaining 10\% for testing the image-expression alignment effect of ExpCLIP.

\subsubsection{BEAT}
We use BEAT \cite{liu2022beat} to train the speech-driven facial animation generator. BEAT comprises 76 hours of speech data, paired with 52D facial blendshape weights. The dataset is collected from 30 speakers, who perform in 8 distinct emotional styles and across 4 different languages. For our experiments, we exclusively employ speech data from English speakers, which totals approximately 35 hours.

\subsection{Implementation Details}
Our framework is implemented by Pytorch\cite{paszke2019pytorch}. 
For ExpCLIP, we train a transformer auto-encoder \cite{vaswani2017attention} with 8 layers for both the encoder $\mathcal{E}$ and $\mathcal{D}$. The text encoder and image encoder from \emph{CLIP-ViT-B/32} are utilized. We set the values of $\lambda_1=1, \lambda_2=\lambda_3=10$. ExpCLIP is trained with a learning rate of 1e-5 and a batch size of 256.
For the animation generator, we employ a 4-layer transformer encoder as the self-attention pooling module. An 8-layer transformer decoder is used to modulate the speech features and style embeddings. The animation decoder consists of 2-layer fully connected layers. Each training sample has a duration of 64 frames with FPS=15. The entire framework is trained using the Adam optimizer \cite{kingma2014adam} on a single A100 GPU.

\subsection{ExpCLIP}

\subsubsection{Multi-modal alignment}

ExpCLIP aligns expressions, text, and images into a joint embedding space. We examine this capability through three types of tasks: text-to-expression, text-to-image, and image-to-expression.

Text-to-expression transforms text to blendshape weights, which are rendered to meshes for better visualization. In-domain text from the test set and out-of-domain text  are collected for evaluation. Image-to-expression can be treated as a 3D face reconstruction task, which transforms images to blendshape weights. Figure \ref{fig:text bs} shows that ExpCLIP exhibits the ability to generate subtle and nuanced expressions, accommodating various types of textual input such as emotion tags, AU descriptions, and sentences. Furthermore, ExpCLIP also excels in the task of recovering intricate facial expressions from in-the-wild images.

As for text-to-image, we employ a text-based image retrieval task to evaluate its performance. we extract image embeddings of the test set of MEAD-3D and employ cosine similarity to retrieve the images that are semantically closest to the embedding of the given text. Figure \ref{fig:text img} indicates ExpCLIP achieves a remarkable alignment between textual and visual semantics.

\begin{table}[t]
\centering
\begin{tabular}{lcc}
\toprule
Datasets & TEAD   & BEAT           \\ \midrule
w/o bs aug       & 0.051          & 0.069          \\
w/. bs aug          & \textbf{0.012} & \textbf{0.021} \\ \bottomrule
\end{tabular}
\caption{MSE of blendshape weights reconstruction on the test set of TEAD and BEAT.}
\vspace{-0.2cm}
\label{tab:bs}
\end{table}

\begin{figure}[t]
    \centering
    \includegraphics[width=0.7\linewidth]{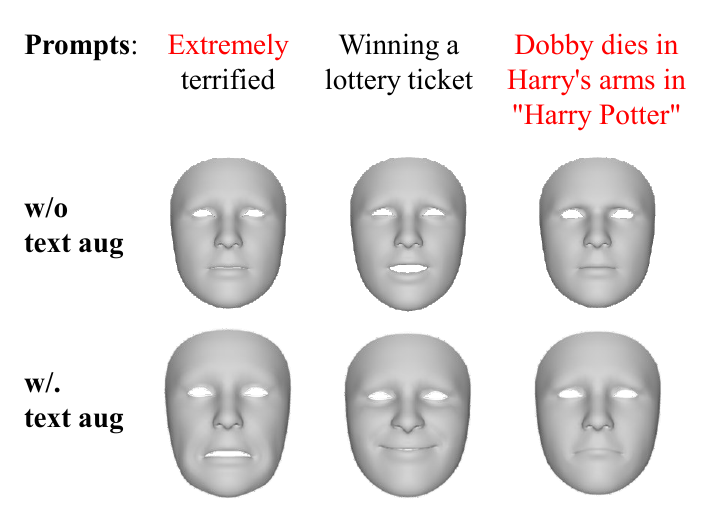}
    \caption{Qualitative results of ablation study for text augmentation. Red words indicate \emph{out-of-domain} texts.}
    \label{fig:text aug}
    \vspace{-0.3cm}
\end{figure}

\subsubsection{Expression manifold smoothness} 

We demonstrate the smoothness of the learned expression manifold in Figure \ref{fig:tsne}. We obtain  embeddings of facial expressions in the test set of TEAD . These embeddings are then projected onto a 2D space using t-SNE \cite{van2008visualizing}. Subsequently, we sample pairs of distant points and interpolate between them. The interpolated values are passed through the decoder of ExpCLIP to reconstruct the corresponding blendshape weights. As observed, it illustrates ExpCLIP achieves  smooth semantic transition between distinct facial expressions.

\subsubsection{Ablation study} 
We conduct an ablation study to validate the effectiveness of blendshape augmentation. The mean squared error (MSE) of blendshape weight reconstruction on the test set of TEAD and BEAT is presented in Table \ref{tab:bs}. The results demonstrate that blendshape augmentation significantly contributes to reducing the reconstruction error of ExpCLIP.
To evaluate the influence of text augmentation, we train a model without text augmentation on TEAD. The qualitative outcomes are depicted in Figure \ref{fig:text aug}. It is evident that text augmentation enhances the consistency between the generated expressions and out-of-domain text prompts.

\begin{figure*}[t]
    \centering
    \includegraphics[width=0.9\linewidth]{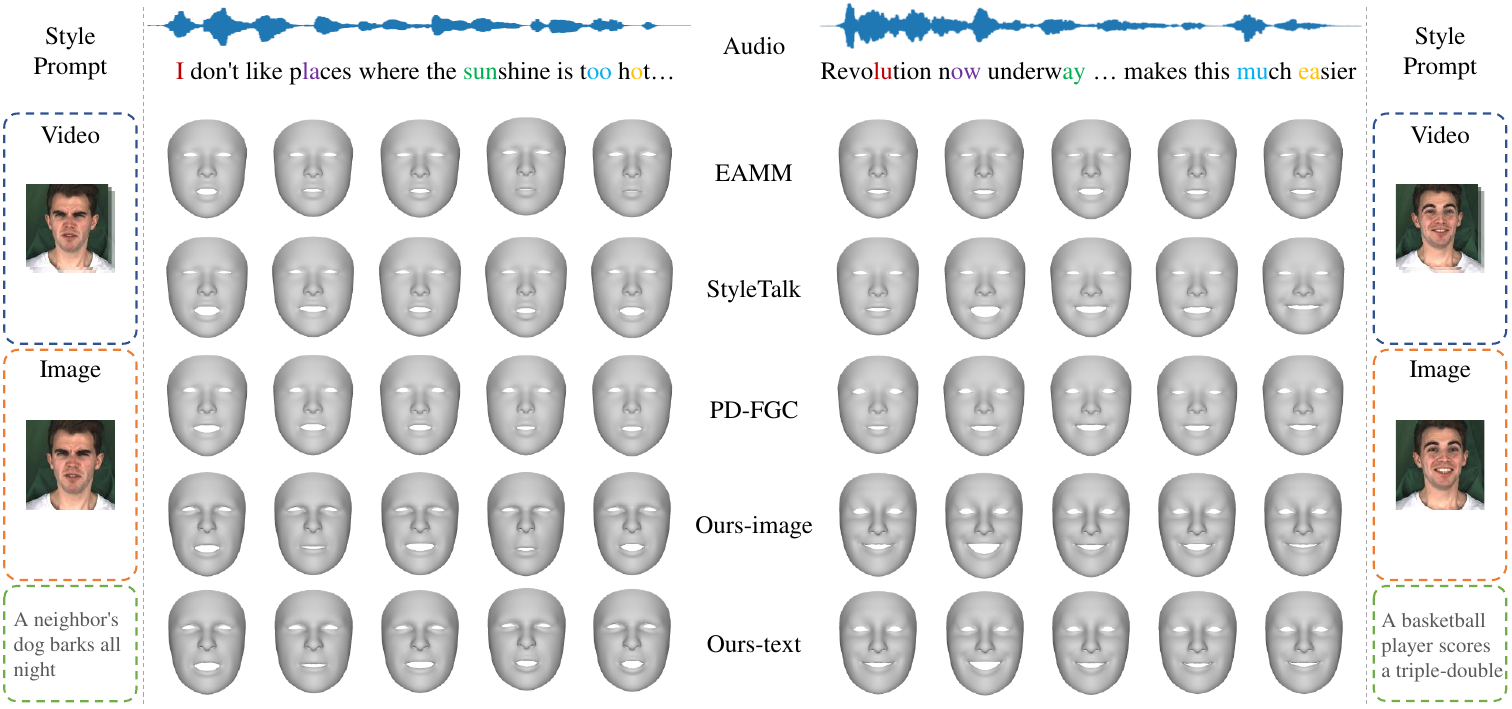}
    \caption{Qualitative comparisons with SOTA stylized talking face generation methods. Note that the speaking style of our method can be guided by text description or emotional image. }
    \label{fig:qualitative}
    \vspace{-0.3cm}
\end{figure*}

\subsection{Emotional speech-driven facial animation}
\subsubsection{Baselines}
As the code for the previous emotion-controllable speech-driven facial animation method has not been released, we compare our method with SOTA emotion-controllable talking face generation methods including: \textbf{EAMM} \cite{ji2022eamm}, \textbf{StyleTalk} \cite{ma2023styletalk} and \textbf{PD-FGC} \cite{wang2023progressive}. We employ the 3D face reconstruction method \cite{guo2022perspective} to reconstruct facial meshes from their generated talking face videos  for comparative analysis. All these methods employ facial images or video templates as emotional prompts. For comparison purposes, we manually annotate textual descriptions for each template.  Notably, as both PD-FGC and our method accept image-based emotional prompts, we can utilize this setting for comparison.

Due to the lack of appropriate quantitative metrics to characterize the precision of emotion control in animation generation, we only conduct qualitative evaluations and user studies to evaluate the above methods.

\subsubsection{Qualitative results}
The qualitative results are illustrated in Figure \ref{fig:qualitative}. The image and video templates are extracted from the test set of MEAD. 
Both BEAT and MEAD provided speech samples for evaluation. 

As can be seen, our method ensures a high level of consistency between the emotional expression in the generated animation and the provided prompt, while also achieving accurate lip synchronization. Even in the more challenging setting of using text as an emotional prompt, our approach achieves high accuracy in emotion control. In comparison, EAMM exhibits inferior lip synchronization and demonstrates significant inconsistency between the emotion of the generated animation and the reference video. Although StyleTalk and PD-FGC achieve satisfactory lip synchronization, their accuracy in emotion control falls short compared to our approach. Additionally, their control methods are inconvenient for users as they require searching for desired emotional reference templates. In contrast, our text-based control approach is flexible and easy to manipulate.


\subsubsection{User study}
We further conduct user studies to evaluate the performance  of the comparative methods. 
We create 10 emotional animations for each method, accompanied by corresponding reference videos or images, as well as text prompts. We invite 20 volunteers to rate these methods from 1 to 5, with higher scores indicating better performance. The volunteers are asked to rate the methods based on the following three aspects: 1. Lip synchronization, 2. Emotion consistency with the reference video/image, and 3. Emotion consistency with the text prompt. 

As shown in Table \ref{tab:user study}, our performance in lip synchronization exceeds that of StyleTalk and PD-FGC, and markedly surpasses EAMM. Regarding emotion consistency with video/image, our text-based control method approximates the level of StyleTalk and PD-FGC, while our image-based control method exhibits superior performance compared to other methods. As for emotion consistency with text, our text-based control method outperforms all other methods, and our image-based control method also surpasses other video or image-based approaches. The above results demonstrate that our method not only achieves precise lip synchronization but also enables accurate emotion control.

\renewcommand{\arraystretch}{1.2} 
\begin{table}[t]
\centering
\resizebox{0.5\textwidth}{!}{%
\begin{tabular}{cccccc}
\toprule
\textbf{Methods}                                                                & \textbf{EAMM}         & \textbf{StyleTalk} & \textbf{PD-FGC} & \textbf{Ours-image}    & \textbf{Ours-text}       \\ \midrule
\begin{tabular}[c]{@{}c@{}}Lip sync\\  \end{tabular}                            & 2.01                  & 3.55               & 3.49         & 3.59   & \textbf{3.63}          \\ 
\begin{tabular}[c]{@{}c@{}}Emotion consistency \\ with video/image\end{tabular} & 2.11                  & 3.63               & 3.55            & \textbf{3.72}   & 3.65        \\ 
\begin{tabular}[c]{@{}c@{}}Emotion consistency \\ with text\end{tabular}        & 1.99                  & 3.48               & 3.37       & 3.75      & \textbf{3.93}          \\ \bottomrule

\end{tabular}
}
\caption{Results of the user study}

\label{tab:user study}
\vspace{-0.3cm}
\end{table}

\begin{figure}[t]
\begin{center}
\includegraphics[width=0.9\linewidth]{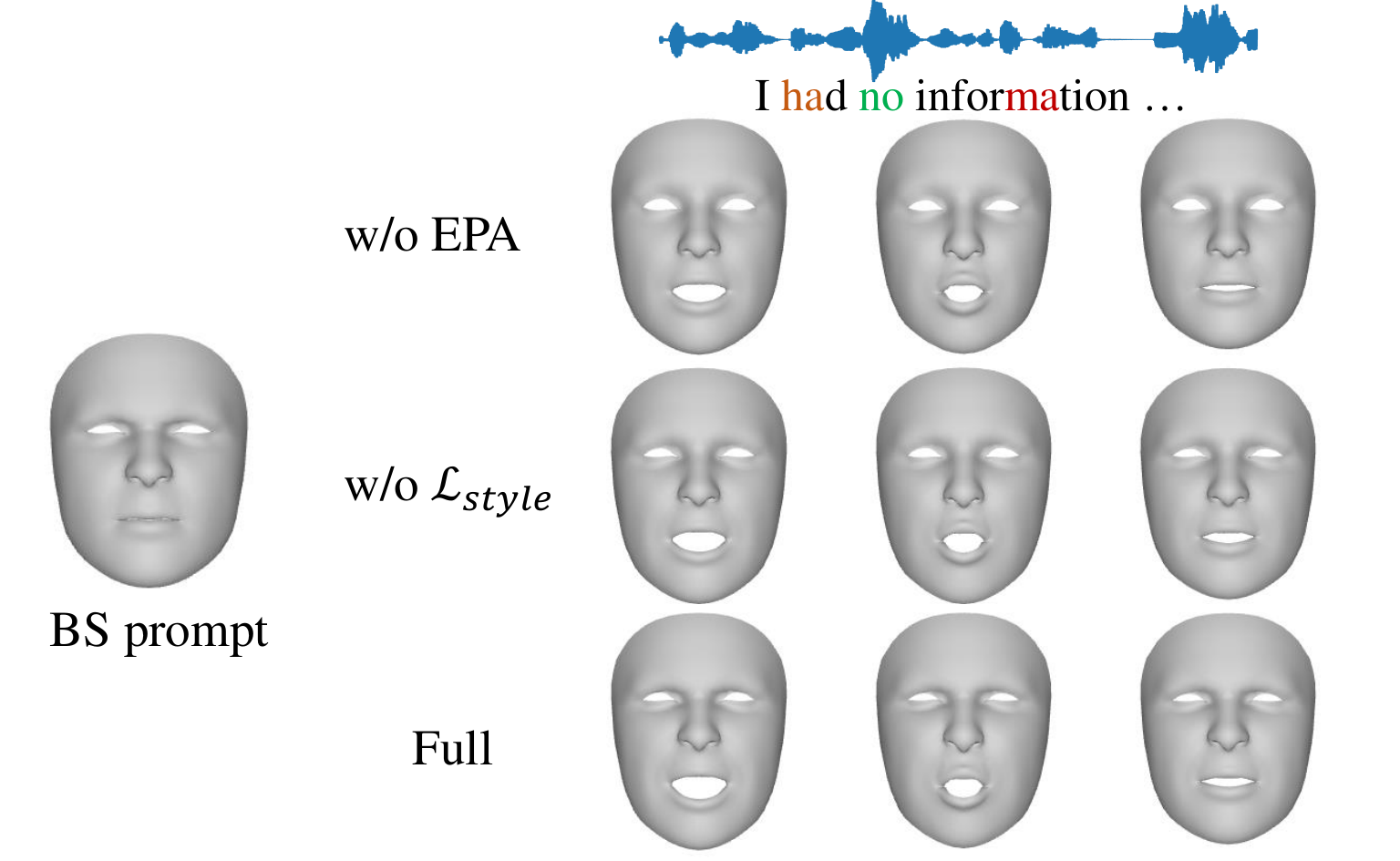}
\end{center}
\caption{Ablation studies of EPA and style loss.}
\label{fig:user_study}
\vspace{-0.3cm}
\end{figure}

\subsubsection{Ablation study}
We conduct ablation studies to validate the effect of EPA and the proposed style loss
$\mathcal{L}_{style}$, the qualitative results are shown in Figure \ref{fig:user_study}. Notably, to better visualize the differences between the comparisons, we utilize blendshape weights as the prompt, rather than text. As illustrated, when there is no EPA, if an unseen emotion is input, the resulting animation exhibits minimal expression. This is due to the limited generalization ability of the model toward unfamiliar emotions. However, the integration of EPA markedly enhances the emotional expressiveness of the animation. Despite this improvement, the consistency with the expression prompt remains somewhat less than ideal. The incorporation of style loss $\mathcal{L}_{style}$ further intensifies the emotional impact.

\section{Conclusion}
This paper introduces, for the first time, text-guided emotional speech-driven facial animation. To achieve this, a large-scale text-expression dataset TEAD is proposed, and ExpCLIP is trained on this dataset to align features of text and expressions. Experimental results demonstrate that the proposed framework achieves high accuracy and flexibility in emotional-controlled animation generation.


\end{document}